\documentclass[runningheads]{llncs}
\usepackage{amssymb}
\usepackage{tablefootnote}
\usepackage{graphicx}
\usepackage{todonotes}
\usepackage{booktabs}
\usepackage{multirow}
\usepackage{hyperref}
\usepackage{subcaption}
\usepackage[export]{adjustbox}

\usepackage[utf8]{inputenc}
\newcommand{\block}[1]{%
  \raisebox{\dimexpr(\fontcharht\font`X-1em)/2}{\rule{1em}{#1\dimexpr1em/8}}%
}
\DeclareUnicodeCharacter{2581}{\block{1}}
\usepackage{pifont}
\begin{document}

\title{Revisiting N-Gram Models: Their Impact in Modern Neural Networks for Handwritten Text Recognition}
\titlerunning{Revisiting N-Gram Models}

\author{Solène Tarride\orcidID{0000-0001-6174-9865} \and
Christopher Kermorvant\orcidID{0000-0002-7508-4080} 
}
%
\institute{TEKLIA, Paris, France}

\maketitle              

\begin{abstract}
In recent advances in automatic text recognition (ATR), deep neural networks have demonstrated the ability to implicitly capture language statistics, potentially reducing the need for traditional language models. This study directly addresses whether explicit language models, specifically n-gram models, still contribute to the performance of state-of-the-art deep learning architectures in the field of handwriting recognition. We evaluate two prominent neural network architectures, PyLaia \cite{laia} and DAN \cite{dan}, with and without the integration of explicit n-gram language models. Our experiments on three datasets - IAM \cite{IAM-database}, RIMES \cite{RIMES-database}, and NorHand v2 \cite{norhand-v2} - at both line and page level, investigate optimal parameters for n-gram models, including their order, weight, smoothing methods and tokenization level. The results show that incorporating character or subword n-gram models significantly improves the performance of ATR models on all datasets, challenging the notion that deep learning models alone are sufficient for optimal performance. In particular, the combination of DAN with a character language model outperforms current benchmarks, confirming the value of hybrid approaches in modern document analysis systems.



\end{abstract}

\section{Introduction}

Before the era of deep neural networks, handwriting recognition systems \cite{Kozielski2013a,Bluche2013c,Arora2019}, derived from automatic speech recognition, usually combined an optical model, designed to generate character hypotheses from the image, and a language model, responsible for reevaluating these hypotheses using language statistics. The introduction of the statistical language model had significantly reduced the transcription error rates. 

As deep neural network performance drastically improved, the need for statistical language models was less prevalent. 
The rise of transformers and their implicit language modeling capacities \cite{dan,trOCR,nougat} eclipsed even more the need for explicit language models for Automatic Text Recognition (ATR). 
Today, a key research area consists in improving models by integrating Large Language Models (LLMs) implicitly into ATR \cite{doc-llm} or Automatic Speech Recognition (ASR) \cite{asr-llm} models, as LLMs have demonstrated impressive modeling capacities.

However, transformers and LLMs require extensive training data and have primarily shown efficiency in the context of well-resourced, contemporary languages. This efficiency might not readily apply to low-resource languages or historical context. 
In contrast, n-gram models can find applicability in this context, as they are able to capture language statistics even with limited training data. If n-gram models have already demonstrated their utility in enhancing convolutional recurrent models with CTC \cite{laia,rethinking-line-atr}, their impact on transformers remains unexplored. 

 This study aims to fill this gap by investigating how explicit n-gram language models can complement modern neural network architectures to improve ATR performance on diverse linguistic datasets. We explore different strategies for integrating n-gram models and investigate their impact on the efficiency and accuracy of ATR systems.
 
This paper is organized as follows. The next Section \ref{sec:related_works} provides an overview of research related to the topic of language models for ATR. In Section \ref{sec:methodology}, we introduce the datasets used and our methodology. Section \ref{sec:experiments} explores the optimal parameters for language modeling. Finally, results are presented and discussed in Section \ref{sec:results}.  


\section{Related works}
\label{sec:related_works}
In this section, we present research that addresses the significance of language models in improving Automatic Text Recognition (ATR).

\subsection{Tokenization levels in language modeling}

The granularity of units used in language modeling is referred to as the tokenization level. 

Character-level tokenization treats each character as a separate token and is widely employed for ATR \cite{dan,laia,nougat}. Character-level language models are highly valuable for capturing fine-grained patterns, particularly in languages with complex scripts. 

On the other hand, word-level tokenization treats entire words as tokens. 
Word-based language models prove effective in capturing syntactic and semantic relationships in text. Nevertheless, they require substantial training datasets and are prone to out of vocabulary (OOV) issues.

Subword-level tokenization can be viewed as a compromise between the character and word levels, as it involves breaking words into smaller units than words. This level is suited for languages with complex word structures and is also beneficial for handling out-of-vocabulary words. Many tokenization algorithms are available for subword tokenization, including SentencePiece \cite{SentencePiece}, Byte-Pair Encoding (BPE) \cite{sennrich-etal-2016-neural}, and Unigram \cite{kudo-2018-subword}. Subword tokenization is particularly popular in machine translation and text generation tasks, and it has also become increasingly important for some ATR systems \cite{trOCR}.

\subsection{Improving Automatic Text Recognition models with language models}
Language models play a crucial role in improving the performance of Automatic Text Recognition (ATR) systems, especially in the case of noisy documents or ambiguous writing styles.
Language models can help by examining the previous words, subwords, or characters, enabling ATR systems to make more contextually informed decisions. 

\subsubsection{Implicit language modeling with neural networks}
In recent architectures, language modeling is implicitly learned by the decoder. Recurrent Neural Networks handle sequential data by maintaining hidden states that capture context from previously predicted tokens.  However, they suffer from vanishing gradient problems and limited context capture. To address these issues, more advanced architectures, such as Long Short-Term Memory (LSTM) \cite{laia,Voigtlaender} and Gated Recurrent Units (GRUs) \cite{HTR-Flor}, have been introduced. These models are designed to overcome the vanishing gradient problem and enhance the capture of long-range dependencies, making them particularly effective in scenarios where context over extended sequences is crucial for accurate processing and understanding, for example in ATR \cite{laia}.
Recently, the inclusion of attention mechanisms by Transformer models has led to a significant change in NLP. They can effectively model long sequences and now serve as the basis for top-performing models in various applications, such as ATR \cite{nougat,dan,trOCR}. 

\subsubsection{Explicit language modeling with statistical language models}

Since ATR often encounters challenges on distorted or noisy documents, researchers have also relied on explicit statistical language models to improve the performance of CTC-based neural networks \cite{kumar2017lattice,laia,zhang2020scut}.
The most common form of explicit models used in this context are n-gram models, which are probabilistic models that capture the statistical relationships between sequences of tokens in natural language. They are based on the assumption that the probability of the next token in a sequence depends only on a fixed-sized window of previous tokens. Smoothing strategies are often applied to avoid assigning a zero probability to tokens that have not been previously encountered. These statistical models have proven their worth in other language-related tasks, such as ASR \cite{RNNLM-asr}. 
Explicit language models consider the likelihood of text sequences, and can be used as prior in the decoding process to improve recognition. This context-awareness allows ATR systems to choose more contextually appropriate interpretations, resulting in improvements in recognition. So far, n-gram models have been combined to ATR models at character \cite{rethinking-line-atr,zhang2020scut,n-grams-in-atr} and word \cite{laia,n-grams-in-atr} levels. 

\subsubsection{Error correction as post-processing}

Finally, another way to improve recognition consists in correcting errors as a post-processing step \cite{survey-postOCR}. Different strategies have been proposed over the years, either based on isolated word correction (merging OCR output, lexical approaches, ...) or based on contextual language models (statistical language models, sequence-to-sequence models \cite{soper-etal-2021-bart}). 
If error correction techniques can be helpful during post-processing, it is often preferable to rescore hypotheses during decoding in order to preserve all hypothesis probability distributions from the original ATR model. 

\subsection{Discussion}

After reviewing the literature, two observations can be made.

First, explicit language modeling in ATR has been gradually replaced by models with implicit language modeling capabilities, such as transformers \cite{dan,nougat}. However, the potential integration of explicit n-gram models with transformers remains unexplored, although similar combinations have been investigated in other domains \cite{ngram-swin,roy2022ngrammer}. In their study, Diaz et al. \cite{rethinking-line-atr} found that using CTC-based decoders with explicit n-gram models outperformed transformer decoders, and suggested that combining transformer models with n-gram models could further improve ATR results.

Second, ATR models traditionally operate at the character or word level \cite{laia}. However, there is a current trend to explore the integration of intermediate tokenization levels, such as subwords \cite{trOCR,SentencePiece} or multiple tokenization levels \cite{n-grams-in-atr}. The integration of explicit language models trained at different text granularities could improve ATR systems, traditionally working at character-level.

\section{Datasets and models}
\label{sec:methodology}
In this section, we describe the datasets and models used for our experiments. 
\subsection{Datasets}
\begin{figure}[ht]
    \centering
    \begin{subfigure}[b]{0.25\textwidth}
         \includegraphics[width=0.9\textwidth, center]{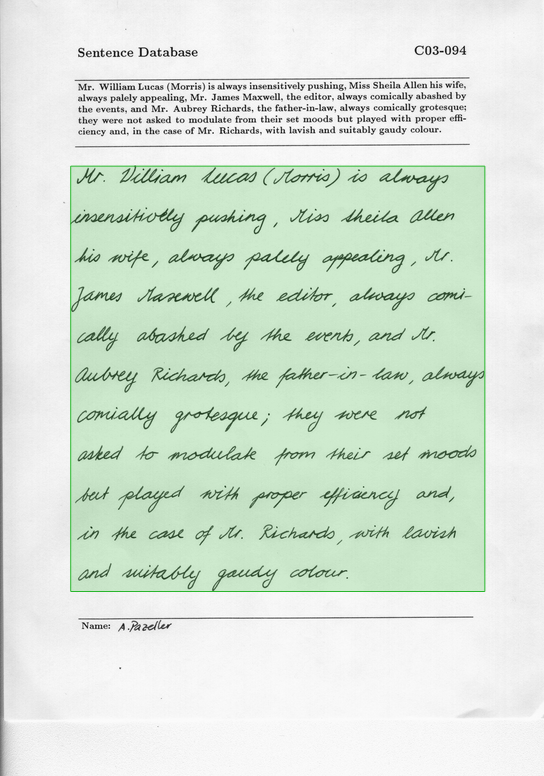}
         \caption{IAM \cite{IAM-database}}
         \label{fig:iam}
     \end{subfigure}     
    \begin{subfigure}[b]{0.25\textwidth}
         \includegraphics[width=0.9\textwidth, center]{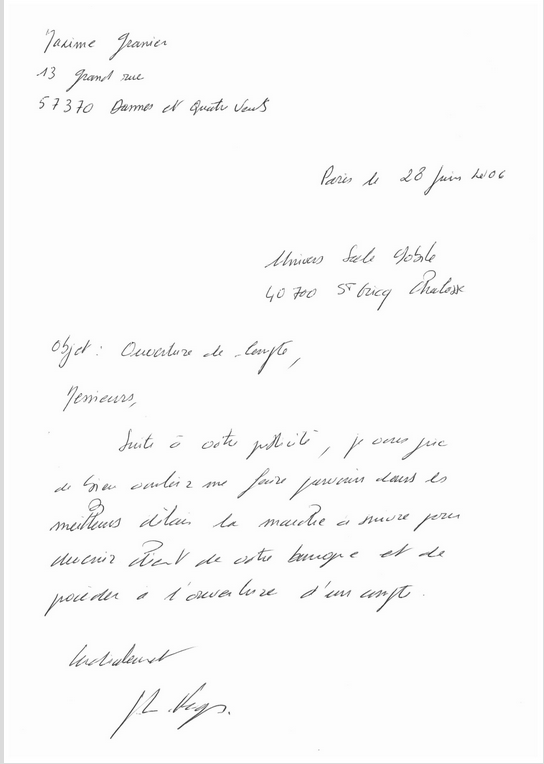}
         \caption{RIMES \cite{RIMES-database}}
         \label{fig:rimes}
     \end{subfigure}   
    \begin{subfigure}[b]{0.47\textwidth}
         \includegraphics[width=0.9\textwidth, center]{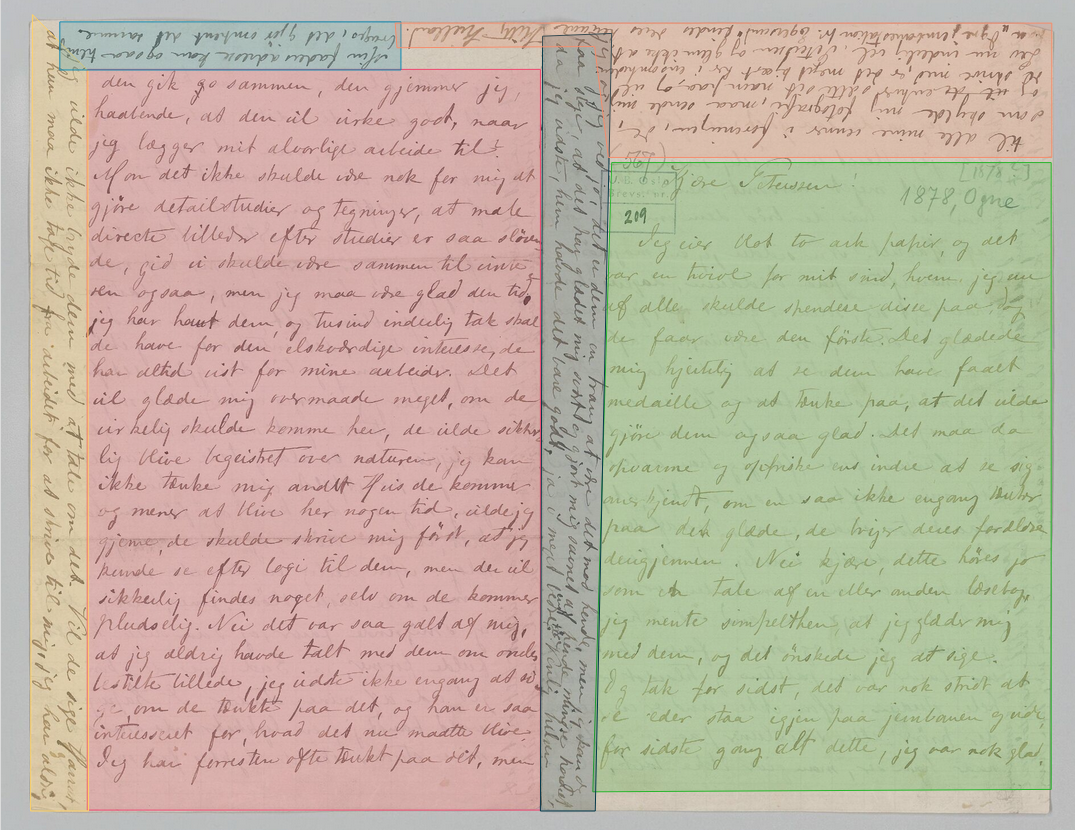}
         \caption{NorHand v2 \cite{norhand-v2}}
         \label{fig:humu}
     \end{subfigure}   
    \caption{Examples of pages from the three datasets used in this work. Note that we use full pages for RIMES and paragraphs for IAM (to exclude the printed header) and NorHand v2 (to simplify reading order). Paragraphs are highlighted.}
    \label{fig:datasets}
\end{figure}

We evaluate our systems on three datasets, each covering a different language: IAM, RIMES and NorHand. Figure \ref{fig:datasets} presents an example of each dataset, and Table \ref{tab:splits} details the dataset splits.

\addtolength{\tabcolsep}{4pt}    
\begin{table}[ht]
    \centering
    \caption{Number of lines, paragraphs and pages used for training, validation and testing for each dataset.}
    \label{tab:splits}
    \begin{tabular}{l|rrr|rrr|rrr}
    \toprule
        \textbf{Dataset} & \multicolumn{3}{c|}{\textbf{Lines}} & \multicolumn{3}{c|}{\textbf{Paragraphs}} & \multicolumn{3}{c}{\textbf{Pages}}  \\
         & \textbf{Train} & \textbf{Val} & \textbf{Test} & \textbf{Train} & \textbf{Val} & \textbf{Test} & \textbf{Train} & \textbf{Val} & \textbf{Test}  \\
        \midrule
        IAM & 6,482 & 976 & 2,915 & 747 & 116 & 336 & 747 & 116 & 336 \\
        RIMES & 10,195 & 1,138 & 778 & - & - & - & 3,700 & 249 & 100 \\
        NorHand & 146,422 & 15,089 & 1,573  & 8,666 & 885 & 90 & 6,551 & 708 & 126 \\
    \bottomrule 
    \end{tabular}
\end{table}
\addtolength{\tabcolsep}{1pt}

\subsubsection{IAM}
The IAM dataset \cite{IAM-database} is composed of modern documents in English, written by 500 writers. It includes 747 training pages with corresponding transcriptions. For this dataset, volunteers were asked to write a short text extracted from a book. 
For our experiments, we use the RWTH split. We perform experiments at two levels: text lines and paragraphs. 
Since our focus is on handwriting recognition, we use paragraphs instead of pages to exclude the printed instructions in the header.

\subsubsection{RIMES}
The RIMES dataset \cite{RIMES-database} is composed of administrative documents written in French. We use the \textit{Letters} subset that includes 3700 training pages with corresponding transcriptions. For this subset, volunteers were instructed to write an administrative letter in their own words. 
For lines, we use the RIMES-2011 version which includes only text lines from the letters' body (address, reference numbers, signatures are excluded). 

\subsubsection{NorHand v2}
The NorHand \cite{HuMu-opensourceHTR} dataset consists of Norwegian letters from the 19th and early 20th century. For the experiments, we train models on the version 2 of the NorHand dataset\footnote{\url{https://zenodo.org/records/10555698}} \cite{norhand-v2}, recently released on Zenodo. On this dataset, the test set is designed to maximize diversity, with each test page written by a different author. 
Note that the paragraphs on a page may have varying orientations, as illustrated in Figure \ref{fig:humu}. As a result, various reading orders can appear within a single page. We decided to train our model on paragraphs instead of pages to ensure that the reading order is consistent in a single image. 
Since we have no information on the orientation, we include all paragraphs in the training and validation phases for paragraph-level training. However, during evaluation, we restrict our assessment to paragraphs with the expected reading order.
For line-level training, we include all horizontal lines, which does not prevent us from training on upside-down text-lines.

\subsection{Models}
Our experiments are carried out using two models: PyLaia \cite{laia} and DAN \cite{dan}. 
These models were chosen because they are open-source, and yield state-of-the-art results at line-level \cite{HuMu-opensourceHTR} and page-level \cite{dan,key-value}.

\subsubsection{PyLaia}\footnote{\url{https://github.com/jpuigcerver/PyLaia}} is an open source model for handwritten text recognition. It combines 4 convolutional layers and 3 recurrent layers, and is trained with the CTC loss function. The last layer is a linear layer with a softmax activation function that computes probabilities associated with each character of the vocabulary. 

PyLaia is trained on text line images resized with a fixed height of 128 pixels and a batch size of 8. We use early stopping to avoid overfitting: the training is stopped after 80 epochs without improvement. PyLaia can be trained in a few hours on a NVIDIA GeForce RTX 3080 Ti (12Go), depending on the dataset size.

\subsubsection{DAN}\footnote{\url{https://github.com/FactoDeepLearning/DAN}} is an open source hybrid model, consisting of a CNN encoder folllowed by an attention-based Transformer decoder. It is designed for handwritten text recognition, and can work directly on images of paragraphs or pages. DAN is trained with the cross-entropy loss function. The last layer is a linear layer with a softmax activation function that computes probabilities associated with each character of the vocabulary. 

For each dataset, we train DAN on text lines and on pages (or paragraphs) to study the impact of language models at different scales. When training on lines, we resize images to a fixed height of 128 pixels, and use a batch size of 8. When training on paragraphs or pages, we use a batch size of 2 due to memory constraints, and rescale images so that they do not exceed 1250 pixels x 2500 pixels. DAN can be trained in a few days on a NVIDIA TESLA V100 (32Go), depending on the dataset size.

\subsection{Explicit language modeling}
In this section, we describe how language models can be built and combined with PyLaia and DAN.

\subsubsection{Building n-gram language models}

Two prominent libraries for constructing n-gram language models are:

KenLM is a widely-used and efficient library for creating and using n-gram language models. It offers tools for training models on large text corpora and features fast and memory-efficient decoding. KenLM supports various tokenization levels, making it versatile for different ATR and ASR applications.

The SRILM toolkit is a comprehensive solution for language modeling, including n-gram models. SRILM provides a range of utilities for building, training, and applying language models. It is extensively used in academia and industry for ASR, machine translation, and other language processing tasks, offering flexibility and robust performance.
Both toolkits are compatible with Flashlight/Torchaudio CTCDecoder.
%
%

\subsubsection{Language model integration}

Once trained, PyLaia and DAN are used to decode with and without an explicit language model.

Decoding without any language model consists in a greedy decoding, where the character with the highest probability is selected at each time step. For DAN, the decoding stops when the end-of-sequence token is predicted (seq2seq model), while for PyLaia there is no such token (CTC model).

In contrast, decoding with a language model relies on the beam search decoding algorithm. This method combines the probability matrix from the optical model with the conditional probabilities of the language model, taking into account lexicon constraints.
In our implementation, language models are built at character, subword and word levels using the KenLM library. 
We rely on torchaudio's \texttt{ctc\_decoder}\footnote{\url{https://pytorch.org/audio/main/generated/torchaudio.models.decoder.ctc\_decoder.html}}. 

Note that for now, torchaudio supports the Connectionist Temporal Classification (\texttt{CTC}) mode, however the sequence-to-sequence (\texttt{S2S}) mode is not available in Python bindings yet. As a result, the integration in PyLaia was straighforward, but more challenging with DAN. 
To make DAN compatible with the \texttt{CTC} mode, we had to add fake CTC frames between frames to avoid the removal of duplicate characters. This simple trick allowed the use of language models with DAN without impacting the optimal decoding path.
Note that in both case, the optical part of the decoding can be performed on GPU. However, the language model re-scoring is always performed on CPU.

Our implementation of PyLaia combined with n-gram models is now open-source\footnote{\url{https://gitlab.teklia.com/atr/pylaia}} and is described in \cite{pylaia-lib}. 

\section{Language model optimization}
\label{sec:experiments}
In this section, we explore the optimal parameters to use for language modeling. More specifically, we explore the tokenization algorithms, the n-gram order, the weight of the language model, smoothing, and the impact of using additional text data. Optimal parameters are optimized on the NorHand dataset. 

\subsection{Tokenization level}
N-gram language models can be built at different granularities. Representing language as a sequence of words provides a broader context, but presents challenges in dealing with unseen words. In contrast, character modeling eliminates the problem of out-of-vocabulary words, but provides a narrow context because the model only considers the preceding $n$ characters. Choosing subword modeling is a useful compromise that breaks down rare words into more common subwords, reducing the impact of the out-of-vocabulary problem while maintaining a broader context. Since PyLaia and DAN predict sequences of characters, it may be interesting to combine them with n-gram models at a different level of granularity to enrich the language context.

Three levels of tokenization are explored for language modeling: character-level, subword-level, and word-level. 
An example of tokenization at different levels is presented in Table \ref{tab:tokenization}. 

\begin{table}[ht]
    \centering
    \caption{An example of text tokenized at different levels}
    \label{tab:tokenization}
    \begin{tabular}{l|l}
        \toprule
        \textbf{Level} & \textbf{Tokenized text} \\
        \midrule
         Characters &  \texttt{T h e ▁ n u m e r i c a l l y ▁ l a r g e s t ▁ g r o u p}\\
         Subwords & \texttt{The ▁ numer ic ally ▁ large st ▁ gro up} \\
         Words & \texttt{The ▁ numerically ▁ largest ▁ group} \\
         \bottomrule
    \end{tabular}
\end{table}

Character tokenization is straightforward, as each token represents a character.  For word tokenization, we rely on the \texttt{wordpunct\_tokenize} function from the \texttt{NLTK} Python Package, which considers punctuation as separate words. Subword tokenization can be achieved with different tokenization algorithms. In this study, we trained SentencePiece \cite{SentencePiece} with a vocabulary size of 1000 subwords. This value was chosen as a compromise between the number of distinct characters and the number of distinct words. However, further experiments could be performed to determine the optimal value for the vocabulary size.
Two different representations were initially tested for subword tokenization.
\begin{itemize}
    \item \textbf{SentencePiece representation}. The first representation follows the SentencePiece mode, and includes spaces in tokens, e.g. \texttt{▁numer ic ally}. In this case, a single subword has two distinct representations based on whether it appears at the beginning of a word or not.
    \item \textbf{SeparateSpaces representation}. The second representation handles spaces as separate tokens, e.g. \texttt{▁ numer ic ally}. In this case, each subword has a unique representation that does not depend on its position in a word.
\end{itemize}
Our initial experiments demonstrated a marginal improvement in results (1\% WER on Norhand) with the second representation method. Consequently, we adopted this method for future experiments.

\subsection{Order}
N-gram models compute the probability of a token based on the preceding sequence of n-1 tokens. In our evaluation, we compared various values for the order $n$, spanning from 1 to 6 since SRILM and KenLM do not support higher orders at the moment.

Our experiments show that the optimal order depends on the tokenization level of the language model. 
Notably, for character and subword language models, performance consistently improved with the increase in order. Consequently, the peak performance was observed at $n=6$, suggesting that further increases in order would likely lead to continued performance improvement.
In contrast, the best performance was obtained with $n=3$ for word-based language models. 


\subsection{Language weight}

When combining PyLaia and DAN with n-gram language models, the weight assigned to the language model can be defined.

A one-dimensional grid search, spanning from 0 to 5 with a step of 0.5, was applied to determine the optimal weight to give language modeling. Our experiments show that the optimal weight depends on the tokenization level of the language model. 
For character-level and subword-level language models, the most effective weight was determined to be 1.5 for both DAN and PyLaia. On the other hand, the optimal weight for word-level language models was found to be lower, equal to 0.5 for both DAN and PyLaia.


\subsection{Smoothing}

To keep the language model from assigning zero probability to unseen tokens, it is recommended to apply smoothing on the probability distribution \cite{nlp-course}. We conducted experiments to determine the best smoothing method among no smoothing, Kneser-Ney smoothing, and Witten-Bell smoothing.

Our experiments show that Kneser-Ney smoothing \cite{smoothingLM} leads to the best performance by a thin margin. Although it does not have a huge impact on performance, it avoids errors on unseen sequences. Note that the Kneser-Ney smoothing is the default smoothing method used in KenLM.


\section{Results and discussion}
\label{sec:results}
In this section, we present results with and without language models on three datasets, using the parameters described in the previous section. The results are presented in Table \ref{tab:perf-line} at line-level and in Table \ref{tab:perf-page} at page-level. We analyze and discuss the conclusions that can be drawn from our experiments. 

\begin{table}[]
    \centering
    \caption{Comparison of models trained at line-level with different tokenization levels for language modeling. Evaluation results are presented on the test set (\%). Best results for each dataset appear in bold, best results for each model are underlined.}
    \begin{tabular}{l|l|rr|rr|r}
    \toprule
     \textbf{Dataset} &  \textbf{LM level}  & \multicolumn{2}{c}{\textbf{PyLaia}} &  \multicolumn{2}{c|}{\textbf{DAN}} & \textbf{N lines}\\
     & & \textbf{CER} & \textbf{WER} & \textbf{CER} & \textbf{WER} &  \\
    \midrule
    NorHand & None  & 9.72 & 27.78 & 7.85 & 	22.07 & 1494 \\
            & Character &  	\underline{8.23} & 21.65 & \underline{\textbf{7.23}} & \underline{\textbf{19.87}} & 1494 \\
            & Subword & 8.48 & \underline{21.60} & 7.66 & 21.16 & 1494 \\
            & Word    & 8.79 & 23.20 & 9.00 & 27.02 & 1494 \\
            \midrule
    RIMES   & None   & 4.57 & 14.72 & 3.22 & 10.69 & 778 \\
            & Character    & \underline{3.79} & \underline{10.94}  & 3.22 & 10.83 & 778 \\
            & Subword & 4.26 & 12.11  & \underline{\textbf{3.15}} & 10.26 &  778\\
            & Word & 4.28 & 12.59 & 3.19 & \underline{\textbf{10.08}}  & 778 \\
       \midrule
    IAM     & None   &  8.44 &  24.51  & 4.86 & 16.39 & 2,915 \\
            & Character  & \underline{7.50} & \underline{20.98} & \underline{\textbf{4.38}} & \underline{\textbf{14.40}} & 2,915 \\
            & Subword & 8.05 & 21.87 & 4.58 & 14.74 & 2,915 \\
            & Word & 12.02 & 27.70  &  7.56 &  21.80 & 2,915\\ 
    \bottomrule
    \end{tabular}
    \label{tab:perf-line}
\end{table}

\begin{table}[]
    \centering
    \caption{Comparison of models trained at page or paragraph-level with different tokenization levels for language modeling. Evaluation results are presented on the test set (\%). Best results for each dataset appear in bold.}
    \begin{tabular}{l|l|rr|r}
    \toprule
     \textbf{Dataset} &  \textbf{LM level}  & \multicolumn{2}{c|}{\textbf{DAN}}  &  \textbf{N pages}  \\
    & & \textbf{CER} & \textbf{WER} & \\
    \midrule
    NorHand & None  & 12.28 & 28.17 & 81 \\
            & Character & \textbf{11.28} & \textbf{25.04} & 81 \\
            & Subword   & 12.27 & 29.29 & 81 \\
            & Word   &  13.68 &	33.09 &	81\\ 
            \midrule
    RIMES   & None    & 2.22 & 8.23 & 100 \\
            & Character  & \textbf{2.13} & \textbf{7.96} & 100 \\
            & Subword & 2.22 & 8.44 & 100 \\
            & Word & 2.64 & 10.27 & 100 \\
       \midrule
    IAM     & None   &  4.30 & 14.29 & 336 \\
            & Character  & \textbf{3.95} & \textbf{12.71} & 336\\ 
            & Subword  & 4.10 & 13.22 & 336 \\
            & Word & 7.56 &  21.18 & 336\\
    \bottomrule
    \end{tabular}
    \label{tab:perf-page}
\end{table}

\subsection{Impact of language models on performance}
In this section, we explore the effects of tokenization levels on language modeling, investigate the influence of model architecture, and analyze the impact of varying image input levels.

\subsubsection{Tokenization level}
Below, we discuss how different levels of tokenization affect ATR performance across datasets.

On IAM, character-level language model (LM) shows superior performance, as reflected by a small relative reduction of 13\% in word error rate compared to the model without any LM. While subword-based LM contributes marginal improvements, word-based LM tends to show a decline, particularly on IAM, where the word error rate increased by 31\% compared to the model without LM. This trend may be due to the relatively smaller size of the dataset, resulting in lower word frequencies compared to NorHand and Rimes.

On RIMES, the character LM proves to be highly effective for PyLaia, with a relative improvement of 26\% in terms of word error rates, while its impact on DAN Line and DAN Page is more marginal due to already low error rates. Subword language models show moderate efficiency for PyLaia, with minimal impact on DAN. Surprisingly, word language models show a significant decrease for DAN Page, while performing well for DAN Line and PyLaia. 

On NorHand, character-based language modeling emerges as the optimal choice for all models, with an average relative improvement of 15\% in terms of WER. The subword language model is also a solid choice for PyLaia, although it shows a marginal improvement for DAN. The word-based language model is ineffective for DAN and completely negative for PyLaia. 

Overall, character-based language models are the optimal choice, consistently outperforming subword models. Word-based models, on the other hand, lead to a decline in performance, mainly due to challenges with out-of-vocabulary words. 

\subsubsection{Model architecture}

DAN Line outperforms PyLaia on the IAM and RIMES datasets, but surprisingly, PyLaia is better on the NorHand dataset. After analyzing the results, we observe some hallucinations from DAN on NorHand, which is a known issue with Transformer models \cite{dan,nougat}, causing a decline in performance. 

The results show that explicit language modeling improves both PyLaia and DAN. If the relative improvement is comparable on IAM, our experiments indicate that PyLaia benefits more from explicit language modeling than DAN on NorHand, and even more on RIMES. 
Character-based language modeling remained the optimal choice for both models, while word-based language modeling generally failed to improve results. 

\subsubsection{Image level}

DAN Page and Line showed similar trends. Character-based language models were consistently optimal, while sub-word based language models showed limited positive effects. The use of Word LM consistently resulted in a decline in performance. The improvement on RIMES is low, since error rates without language models are already low. However, on IAM and Norhand, explicit language modeling significantly improved results.
It is interesting to note that DAN performs generally better when trained on pages rather than lines, at least for IAM and RIMES. This observation suggests that it benefits from a larger context to predict characters. Surprisingly, this is not the case for NorHand v2, which can be explained by the difficulties of this dataset: variying paragraph sizes (a paragraph can be a set of two lines or a full page) and unknown reading order (trained with different orientations which can impact feature extraction).

\subsection{Impact of language models on speed}
\label{sec:speed}

During inference, computations can be efficiently performed on the GPU up to the language model decoder, which due to current limitations is performed exclusively on the CPU. Consequently, the inclusion of a language model during decoding improves recognition accuracy, but at the cost of a slowdown in prediction speed. Specifically, for PyLaia, decoding with a language model results in a speed reduction by a factor of 10. On the other hand, in the case of DAN, where the decoding time is inherently higher, the proportional impact is less pronounced, as detailed in the table \ref{tab:speed}. We recommend the use of language models for batch processing of documents, since their application may not be suitable for real-time processing.

\begin{table}[]
    \centering
    \caption{Impact of language models on decoding time (in seconds/image).}
    \label{tab:speed}
    \begin{tabular}{l|l|ccc}
    \toprule
    \textbf{Device} & \textbf{LM level} & \textbf{PyLaia Line} &  \textbf{DAN Line} & \textbf{DAN Page} \\
    \midrule
    GPU & None  & \textbf{0.01} & \textbf{0.47} & \textbf{5.72} \\
    GPU + CPU & Character & 0.12 & 0.69 & 8.13 \\
    GPU + CPU & Subword   & 0.09 & 0.67 & 7.78 \\
    GPU + CPU &  Word    & 0.11 & 0.53 & 6.84 \\
    \bottomrule
    \end{tabular}
\end{table}

\subsection{Comparison with state-of-the-art models}

In this section, we compare the results obtained in this study with state-of-the-art models. Results are presented in Table \ref{tab:bench-line} for models trained on text-lines and in Table \ref{tab:bench-page} for models trained on paragraphs or pages. Several key observations can be made. 

First, we establish a new baseline on the NorHand v2 dataset at line-level and page-level. As this dataset was released recently, we could not find any existing models for comparison. On this dataset, DAN systematically outperforms PyLaia, especially when combined with a character language model. 

At page-level, the DAN with language modeling demonstrates state-of-the-art performance on RIMES and IAM. However, at line-level, the same model falls slightly below other methods, while achieving near state-of-the-art results. On these two datasets, the best model is the \textit{S-Attn/CTC}, which uses a character-level language model and relies on additional data during training.  

\begin{table}[]
    \centering
    \caption{Comparison with state-of-the-art models working at line-level. Models introduced in this article appear in bold. Best scores also appear in bold.}
    \begin{tabular}{l|lll|rr}
    \toprule
    \textbf{Dataset} & \textbf{Model} & \textbf{LM level} & \textbf{Add. data} &  \textbf{CER} & \textbf{WER}  \\
    \midrule
    NorHand v2 & \textbf{PyLaia-LM} & None & No & 8.2 & 21.7  \\
     & \textbf{DAN-LM} & Character  & No & \textbf{7.2} & \textbf{19.9} \\    & \textbf{DAN-LM} & Subword  & No & 7.7 & 21.2  \\
    \midrule           
    RIMES   & SFR \cite{wigington2018} & Character & No & 2.1 & \textbf{9.3} \\
            & S-Attn / CTC  \cite{rethinking-line-atr} & Character & Yes & \textbf{2.0} & - \\
            & \textbf{DAN-LM} & Character  & No & 3.2 & 10.1 \\
            & MDLSTM \cite{Voigtlaender} & Word & No & 2.8 & 9.6 \\
            & Laia \cite{laia} & Word & No & 4.4 & 12.2  \\
    \midrule
    IAM & TrOCR large \cite{trOCR} & None & Yes & 2.9 & - \\
            & VAN \cite{van} & None & No & 5.0 & 16.3 \\
            & S-Attn / CTC \cite{rethinking-line-atr} & Character & Yes & \textbf{2.8} & - \\
            & \textbf{DAN-LM}& Character  & No & 4.4 & 14.4 \\
     & MDLSTM \cite{Voigtlaender} & Hybrid & No & 3.5 & \textbf{9.3} \\

    \bottomrule
    \end{tabular}
    \label{tab:bench-line}
\end{table}
\begin{table}
    \centering
    \caption{Comparison with state-of-the-art models working at page-level (RIMES) or paragraph-level (IAM, NorHand). Models introduced in this article appear in bold. Best scores also appear in bold.}
    \begin{tabular}{l|lll|rr}
    \toprule
    \textbf{Dataset} & \textbf{Model} & \textbf{LM level} & \textbf{Add. data} &  \textbf{CER} & \textbf{WER}  \\
    \midrule
    NorHand v2 & \textbf{DAN} & None & No & 12.3 & 28.2  \\
     & \textbf{DAN-LM} &  Character   & No  & \textbf{11.3} & \textbf{25.0} \\
     & \textbf{DAN-LM} & Subword & No & 12.3 & 29.3 \\
    \midrule           
    RIMES  & Encoder Decoder \cite{bluche2016} & None & No  & 2.9 & 12.6  \\
            & DAN \cite{dan} & None & No &  4.5 & 11.9 \\
            & SFR \cite{wigington2018} & Character & No  & \textbf{2.1} & 9.3 \\
            & \textbf{DAN-LM} & Character  & No  & \textbf{2.1} & \textbf{8.0} \\
    \midrule
    IAM    & SFR \cite{wigington2018} & None & No  & 6.4 & 23.2 \\
            & OrigamiNet \cite{yousef} & None & No & 4.7 & - \\
            & VAN \cite{van} & None & No & 4.5 & 14.6 \\
            & \textbf{DAN-LM} & Character   & No  & \textbf{4.0} & \textbf{12.7} \\
       & Encoder Decoder \cite{bluche2016} & Word & No  & 5.5 & 16.4  \\    
    \bottomrule
    \end{tabular}
    \label{tab:bench-page}
\end{table}

\section{Conclusion}

In this article, we have explored the combination of modern neural networks and n-gram language models. N-gram language models were progressively abandoned, as recent architecture demonstrated impressive language modeling performance. The results of this study highlight that these technologies can be combined to further improve performance in Automatic Text Recognition.
We have studied the impact of language modeling at character, subword, and word levels for two models: PyLaia \cite{pylaia-lib} and DAN \cite{dan}. Since these models predict sequences of characters, our initial intuition was that subword-level language modeling would be optimal, as it would enrich the language context without introducing out-of-vocabulary issues \cite{trOCR}. Our results show that, despite subword language model yielding decent improvement, the optimal performance is achieved using character-level language modeling. 
Finally, we also show that language models improve ATR models whether they are trained on text-lines, paragraphs or full pages. As a result of our study, we set a new standard on IAM, RIMES and NorHand v2 at paragraph and page-level. One of the highlight of this work is that n-gram language models are auto-tuned: they can be trained directly on the training set without requiring any additional data. 

For future works, we plan to explore how different granularities can be combined in language models. Additionally, we would like to study how language models can be leveraged to quickly adapt generic optical models to a specific style or period. 

\section*{Acknowledgement}
This work was supported by the Research Council of Norway through the 328598 IKTPLUSS HuginMunin project.


\bibliographystyle{splncs04}
\bibliography{main}

\end{document}